\documentclass[runningheads]{llncs}

\usepackage{eccv}

\usepackage{eccvabbrv}

\usepackage{graphicx}
\usepackage{makecell}
\usepackage{booktabs}
\usepackage{multirow}
\usepackage{multicol}
\usepackage{tabularx}
\usepackage{enumitem}

\usepackage[accsupp]{axessibility}  

\usepackage{hyperref}

\usepackage{orcidlink}

\AtBeginDocument{%
  \setlength{\textfloatsep}{12pt plus 2pt minus 2pt}%
  \setlength{\intextsep}{10pt plus 2pt minus 2pt}%
  \setlength{\floatsep}{10pt plus 2pt minus 2pt}%
  \setlength{\dbltextfloatsep}{12pt plus 2pt minus 2pt}%
  \setlength{\dblfloatsep}{10pt plus 2pt minus 2pt}%
  \setlength{\abovecaptionskip}{6pt}%
  \setlength{\belowcaptionskip}{0pt}%
}

\begin{document}

\title{Misanthrope: A Privacy-Preserving Keypoint Detector} 


\author{Francesco Vultaggio\inst{1,2}\orcidlink{0009-0005-1622-9816} \and
Predrag Djindjic\inst{3}\orcidlink{0009-0009-8354-0996} \and
Markus Gerke\inst{2}\orcidlink{0000-0002-2221-6182} \and
Sebastian Tschiatschek\inst{3}\orcidlink{0000-0002-2592-0108}\and
Phillipp Fanta-Jende\inst{1}\orcidlink{0000-0001-8733-5425}}

\authorrunning{F.~Vultaggio et al.}

\institute{Austrian Institute of Technology, Assistive \& Autonomous Systems\\
Vienna 1210, Austria. \email{\{name.surname\}@ait.ac.at} \and
Technical University of Braunschweig, Institute of Geodesy and Photogrammetry \\ 
Braunschweig 38106, Germany. \email{\{n.surname\}@tu-bs.de} \and
University of Vienna, Data Mining and Machine Learning \\
Vienna 1090, Austria. \email{\{name.surname\}@univie.ac.at} }

\maketitle

\begin{abstract}
Image matching is a core component of applications such as Simultaneous Localization and Mapping (SLAM), Visual Localization, and Structure from Motion (SfM).  However, the local image features central to this task are vulnerable to inversion attacks, which enable adversaries to reconstruct privacy-sensitive scene content from local features. These attacks pose a particular threat in distributed computing scenarios where the pre-computed features leave edge devices to be processed by remote servers. In this work, we introduce \textit{Misanthrope}, a novel privacy-preserving keypoint detector trained through self-distillation to avoid detecting keypoints on people---a predominant source of privacy-sensitive content in most localization scenarios---thus mitigating inversion attacks at the source rather than through post-hoc obfuscation.
We demonstrate how inverted images from traditional feature detection pipelines can be used to detect and re-identify people in the scene, while Misanthrope is able to mitigate these attacks.
Furthermore, Misanthrope maintains image matching performance on par with the state of the art and even surpasses it in challenging settings where people act as distractors, such as phototourism and in-the-wild odometry.
On the Image Matching Challenge 2021 Phototourism test set, Misanthrope is the top-performing sparse feature extractor in 7 out of 9 scenes. We make our model and its evaluation script available here: \url{https://github.com/fratopa/misanthrope}.

\end{abstract}

\section{Introduction}
\label{sec:intro}

Finding correspondences across images is a fundamental task in computer vision. It sits at the core of many traditional geometric computer vision tasks and downstream applications spanning from Simultaneous Localization and Mapping (SLAM)~\cite{vslam_review,dxslam} to Visual Localization~\cite{vloc_review,vl_review2}, and beyond~\cite{dl_image_matching_review}. 
A traditional image matching pipeline consists of three steps: keypoint detection, keypoint description (i.e., local feature extraction), and finally local feature matching. While the threat to privacy that images pose is apparent and is actively regulated~\cite{gdpr,DataprotectionEU}, modern \textit{inversion attacks}~\cite{invertIMG,invsfm,bits2img,diffinv} have shown that sparse image features similarly constitute a threat to privacy by revealing the contents of the image from which the features have been extracted or the scene in which they have been embedded. A model capable of revealing scene contents from Structure from Motion (SfM) point clouds poses a significant privacy threat to any Visual Localization system reliant on such a map representation~\cite{hloc}.  

The Visual Localization community has responded to these attacks with obfuscation-based strategies that modify the 3D structure of SfM maps to resist inversion while aiming to preserve localization performance~\cite{privacy_permutation,privacy_lines,rayclouds}. However, this line of defense is increasingly fragile: newer inversion methods have demonstrated successful attacks against several of these obfuscated representations~\cite{neighborhood,lines_unprivacy}.

Fundamentally, obfuscation-based approaches treat all scene content as uniformly private, which necessitates non-standard map representations that are incompatible with keypoint-based pipelines~\cite{hloc,vloc_review}, and reduces localization accuracy~\cite{privacy_lines,privacy_permutation,privacy_splat,rayclouds}. This assumption holds in settings such as private residences or industrial facilities, where the scene itself is sensitive. However, it does not hold in general. In many practical deployment settings, such as outdoor urban environments or tourist landmarks, the scene geometry required for localization is inherently public, and privacy-sensitive content is limited to semantically distinct elements within it, most notably people. In these settings, detection and redaction of private data at the image level prior to feature detection is already common practice~\cite{googlestreetviewprivacy}, often as a requirement by law~\cite{gdpr,DataprotectionEU}.

We argue that---where private and localizable content are separable---a more effective strategy is to prevent privacy-sensitive features from being produced in the first place, rather than to obfuscate the entire representation after the fact. Crucially, people also act as visual distractors that degrade matching performance, meaning that a detector that avoids them confers a localization benefit alongside the privacy one. 

However, existing keypoint detectors are not designed to avoid people. Na\"ive attempts to filter keypoints on people using a separate segmentation network are sound but computationally expensive and introduce additional latency, which is particularly problematic for edge devices, where a fixed computational budget means any segmentation network comes at the expense of the feature extractor itself. 

To overcome these shortcomings, we introduce Misanthrope: a keypoint detector trained via a novel, privacy-directed self-distillation~\cite{self_distillation} scheme that avoids detecting keypoints on people. In particular, we leverage DeDoDe~\cite{dedode}---a pre-trained state-of-the-art model for local feature extraction---as a teacher. Its keypoint predictions on the large-scale COCO dataset~\cite{coco} are filtered using semantic person masks, creating a privacy-aware supervisory signal. An identical student network is then trained to replicate this filtered output, teaching it to detect robust keypoints while inherently avoiding people (Figure~\ref{fig:VOCexamples}). This removes the need to deploy a separate segmentation network at inference time, streamlining the pipeline and reducing latency on edge devices compared to techniques that redact private information at the image level before feature extraction.

In this work, we make the following contributions:
\begin{enumerate}[label=\roman*.]
	\item To the best of our knowledge, we provide the first demonstration that individuals can be re-identified from imagery reconstructed via inversion attacks on sparse features.
 	\item We propose Misanthrope, a keypoint detector trained via privacy-directed self-distillation that avoids detecting features on people at inference time, without requiring a separate segmentation module.
 	\item We show that avoiding keypoints on people simultaneously strengthens privacy and improves matching, achieving top results among sparse feature extractors on 7 out of 9 scenes in the Image Matching Challenge 2021 (IMC2021) Phototourism benchmark~\cite{IMC2020}. 
\end{enumerate}

\section{Related Work}
\label{sec:related}
\subsection{Image Matching}
At the core of image matching sits feature extraction. Recent years have seen a shift from hand-crafted feature extraction techniques~\cite{sift,orb} to deep learning-based ones~\cite{dedode,superpoint,aliked,xfeat,disk,roma,loftr}. Many modern strategies~\cite{superpoint,silk,xfeat} use an encoder-dual-decoder architecture, where a shared feature encoder feeds two separate heads for keypoint detection and descriptor extraction. Alternatively, some models unify these tasks into a single output~\cite{disk}, while others, like DeDoDe~\cite{dedode}, employ entirely separate networks for detection and description. The training strategies vary from self-supervision~\cite{superpoint,silk}, to reinforcement learning~\cite{disk}, to SfM-derived labels~\cite{dedode}. A parallel family of \textit{detector-free} matchers~\cite{loftr,roma} bypasses keypoint detection, instead matching dense feature maps directly using the attention mechanism~\cite{attention}. While powerful, these methods are computationally intensive. Critically, their reliance on dense feature maps poses a greater privacy risk, as denser representations provide richer signals for inversion~\cite{invertIMG}. We adopt DeDoDe~\cite{dedode}, whose separate detector and descriptor networks allow training a privacy-aware detector without modifying the descriptor.


\textbf{Semantic guidance}: Prior work has leveraged semantic data in local feature pipelines for two distinct purposes: to enrich descriptors and improve matching quality~\cite{semantic_descriptor, semantic_keypoint}, and to improve robustness to dynamic objects by filtering keypoints detected on them~\cite{slamantic}. The latter is closer to our setting, but is typically achieved via a two-stage pipeline in which a standard detector proposes keypoints and a separate segmentation network generates masks to filter them. Our approach differs in both goal and design: rather than targeting dynamic objects for robustness, we target people for privacy, and we train a single network that inherently avoids detecting keypoints on people, embedding semantic awareness directly into the detector without a separate segmentation stage.

\textbf{Knowledge distillation}: Distillation has been applied in the image matching domain primarily as a model compression technique~\cite{dengModelCompressionHardware2020,Edstedt2025MarDaD,Yao2026JanBinDIst,xfeat}, transferring knowledge from a large feature extractor into a smaller one. We repurpose self-distillation~\cite{self_distillation,bornagain}---where teacher and student share the same architecture---for a fundamentally different end: not to compress a model, but to impose a behavioral constraint on it. Specifically, we use a teacher network to train an architecturally identical student to avoid detecting keypoints on people, embedding a privacy-preserving behavior without altering the model's structure. To the best of our knowledge, this is the first use of self-distillation to embed semantic avoidance directly into a keypoint detector.

\subsection{Privacy Threats and Mitigation Strategies}
The conversion of images into sparse local features does not inherently guarantee anonymization. Recent research on inversion attacks has shown that it is possible to reconstruct detailed images from local features alone~\cite{invertIMG}, to recover entire 3D scenes from SfM models~\cite{invsfm}, simple colorized point clouds~\cite{invPC}, and even the approximate contents associated with a CLIP~\cite{clip} embedding~\cite{diffinv}.  As we demonstrate in our experiments, reconstructed imagery is often of sufficient quality to enable person detection and re-identification, constituting a direct privacy threat.

In response, a significant body of work has focused on mitigating these threats through obfuscation. We can distinguish between two classes of obfuscation: \textit{geometric} obfuscation and \textit{descriptor} obfuscation. Geometric obfuscation techniques aim to modify the derived 3D models to disrupt inversion algorithms. Previously proposed strategies include the permutation of 3D point coordinates~\cite{privacy_permutation} or the use of alternative geometric representations such as lines~\cite{privacy_lines} or ray clouds~\cite{rayclouds}. While effective against specific inversion techniques, geometric obfuscation techniques have been met with the development of more advanced reconstruction algorithms capable of defeating these countermeasures~\cite{neighborhood,lines_unprivacy}, indicating a potential limitation of geometric approaches. Descriptor obfuscation was first explored in~\cite{dusmanuPrivacyPreservingImageFeatures2021,ngNinjaDescContentConcealingVisual2022a,pittalugaLDPFeatImageFeatures2023} and more recently in~\cite{privacy_nerf,privacy_splat}. The first family of techniques aims to modify pre-existing descriptors to make them more resistant to inversion attacks, either via adversarial training~\cite{ngNinjaDescContentConcealingVisual2022a} or by lifting the descriptors to a new subspace with adversarial samples to confuse the reconstruction process~\cite{dusmanuPrivacyPreservingImageFeatures2021,pittalugaLDPFeatImageFeatures2023}. We highlight in particular LDP-Feat~\cite{pittalugaLDPFeatImageFeatures2023}, which both introduces techniques to attack the earlier lifting approach~\cite{dusmanuPrivacyPreservingImageFeatures2021} and proposes a new approach that formally bounds the amount of information any attacker can recover from the obfuscated descriptors. However, the recoverable information is directly proportional to the matchability of the descriptors, meaning that robustness to inversion attacks explicitly comes at the cost of matching utility. More recent works~\cite{privacy_nerf,privacy_splat,pietrantoniSegLocLearningSegmentationBased2023} propose to use segmentation masks to learn private features. Their experiments indicate that segmentation masks can limit the information retrievable from the learned features but cannot fully obfuscate it, and these techniques still come at the cost of poorer matching properties.


This work explores an alternative direction: rather than obfuscating privacy-sensitive data after it has been captured, we prevent it from being produced at detection time. The central insight is that, in many practical settings, privacy-sensitive information can be disentangled from the scene geometry required for localization. Where this separation holds, full scene obfuscation is unnecessary; it suffices to exclude privacy-relevant elements from the feature representation entirely. This principle has a decisive theoretical advantage over obfuscation: information that is absent from the representation cannot be recovered regardless of computational resources or future algorithmic advances, and any residual leakage from imperfect avoidance is measurable and auditable prior to deployment. Moreover, in most instances privacy-sensitive information hinders the image matching process; in these cases, avoiding sampling keypoints on people not only reduces privacy leaks but also improves downstream accuracy. Building on this principle, Misanthrope is trained to inherently avoid detecting keypoints on people, integrating privacy constraints directly into the detection stage without requiring a separate person segmentation module or any post-hoc obfuscation.

\section{Method}
\label{sec:method}
\subsection{Overview}

We propose a simple, yet effective, training procedure for a keypoint detector that inherently avoids detecting features on people, thus mitigating privacy leakage at the source. Our approach, Misanthrope, is based on self-distillation: a student network is trained to mimic a teacher’s predictions while avoiding regions semantically identified as people.

Crucially, the detector architecture remains unchanged, and our approach modifies only the training supervision. This allows any existing detector to be adapted into a privacy-aware variant with minimal implementation changes. In this work we use DeDoDe~\cite{dedode} as the base model: its separate detector and descriptor networks allow us to retrain only the detector while leaving the descriptor untouched. For architectures where the keypoint detector and descriptor heads are attached to a single backbone, such as~\cite{xfeat,superpoint,aliked,disk}, the distillation scheme would need to be extended to fine-tune the detections while maintaining the original descriptors, which is beyond the scope of this work.

\subsection{Teacher-Student Distillation with Semantic Filtering}

We adopt the DeDoDe~\cite{dedode} detector for both teacher and student networks. Let $I$ be an input image and $M$ its corresponding semantic mask, where $M(\mathrm{p}) = 1$ if the pixel $\mathrm{p}$ belongs to a person and $0$ otherwise. Let $T(I;\theta_T) \in \mathbb{R}^{H \times W}$ denote the teacher’s keypoint activation map for the image $I$ given the teacher parameters $\theta_T$. The activation map indicates the likelihood of a pixel being a keypoint.

Our training procedure consists of the following steps:

\begin{enumerate}
    \item \textbf{Teacher prediction:} For each input image $I$, we calculate the teacher's keypoint probability map $P_T(I;\theta_T)=\mathrm{softmax}(T(I;\theta_T))$, where the softmax is taken over all spatial locations, yielding a distribution over pixels.
    \item \textbf{Semantic masking:} We then element-wise zero out the probabilities in regions corresponding to people:
    \[
        P_T'(I;\theta_T;M)(\mathrm{p}) = \begin{cases}
            P_T(I;\theta_T)(\mathrm{p}), & \text{if } M(\mathrm{p}) = 0 \\
            0, & \text{if } M(\mathrm{p}) = 1
        \end{cases}
    \]
    \item \textbf{Re-normalization:} To maintain a valid probability distribution, we rescale the masked $P_T'(I;\theta_T;M)$
    \[
        \hat{P}_T(I;\theta_T;M)(\mathrm{p}) = \frac{P_T'(I;\theta_T;M)(\mathrm{p})}{\sum_{\mathrm{p}'}P_T'(I;\theta_T;M)(\mathrm{p}')}
    \]
    \item \textbf{Student supervision:} The student network with parameters $\theta_S$ is trained to predict a keypoint probability map $P_S(I;\theta_S)$ that replicates $\hat{P}_T(I;\theta_T;M)$ using the Kullback–Leibler (KL) divergence as the loss:
    \[
        \mathcal{L}_{\text{KL}}(\theta_S) = \mathrm{KL}\left( \hat{P}_T(I;\theta_T;M) \,\Vert\, P_S(I; \theta_S) \right)
    \]
\end{enumerate}

The result is a student keypoint detector that learns to assign low activation scores to regions where people appear while preserving robustness in the remaining parts of the image.

\subsection{Training Data}

We use the COCO dataset~\cite{coco} since it has proven to be a reliable dataset for the training of image matching models~\cite{superpoint,silk} and provides person segmentation masks. We filter for images containing at least 1\% of pixels labeled as people, yielding a total of $64{,}115$ training images and $2{,}696$ validation images. A limitation of COCO is that its segmentation masks are derived from polygon annotations and therefore provide only coarse boundaries. However, alternative datasets offering pixel-level semantic masks for people, such as LIP~\cite{gongLookPersonSelfSupervised2017}, are dominated by portraiture-style images in which individuals stand against uniform backgrounds and are untested in the training of image matching models. Other candidate datasets are limited either in the number of images containing people~\cite{cityscapes, pascalVoc} or in scene diversity~\cite{semantic_kitty, cityscapes}.

\subsection{Training}

We implement the distillation procedure in PyTorch using DeDoDe-L~\cite{dedode} as our base model. Both teacher and student networks share the same architecture and pre-trained weights. The teacher model remains frozen throughout training, while only the student is updated.

We train for $60{,}000$ iterations with a batch size of $14$ images using AdamW \cite{adamw} with a weight decay of $10^{-4}$ and Automatic Mixed Precision. We implement a $500$-iteration learning rate warm-up from $10^{-9}$ to $5 \cdot 10^{-4}$. Every $100$ training iterations, we evaluate the KL divergence loss on $15$ batches sampled from a held-out validation split of the filtered COCO dataset, and halve the learning rate if no improvement is observed for $30$ consecutive validation runs. Training was performed on a single NVIDIA RTX 3090\,Ti over a 14-hour period.

\section{Experiments}
\label{sec:experiments}

\subsection{Privacy Preservation}
\label{subsec:privacy_experiments}

To evaluate the ability of Misanthrope to preserve the privacy of individuals depicted in scenes, we perform the following tests. We first measure the model's ability to avoid sampling keypoints on people; to this end, we use the PascalVOC2012 dataset~\cite{pascalVoc}. We then test the extent to which the features extracted by our model can be used to recover privacy-sensitive information using the Person Re-identification in the Wild (PRW) dataset~\cite{prw}, which, unlike most re-identification benchmarks~\cite{duke_reid,market_reid,entireid}, provides full scene images rather than pre-cropped person bounding boxes, allowing us to evaluate the complete pipeline from feature extraction through inversion to detection and re-identification. To invert the features, we use InvSfM~\cite{invsfm}, trained on DeDoDe~\cite{dedode} features extracted from the ImageNet dataset~\cite{imagenet}. 
\begin{figure*}[htbp]
    \centering
    \includegraphics[width=\linewidth]{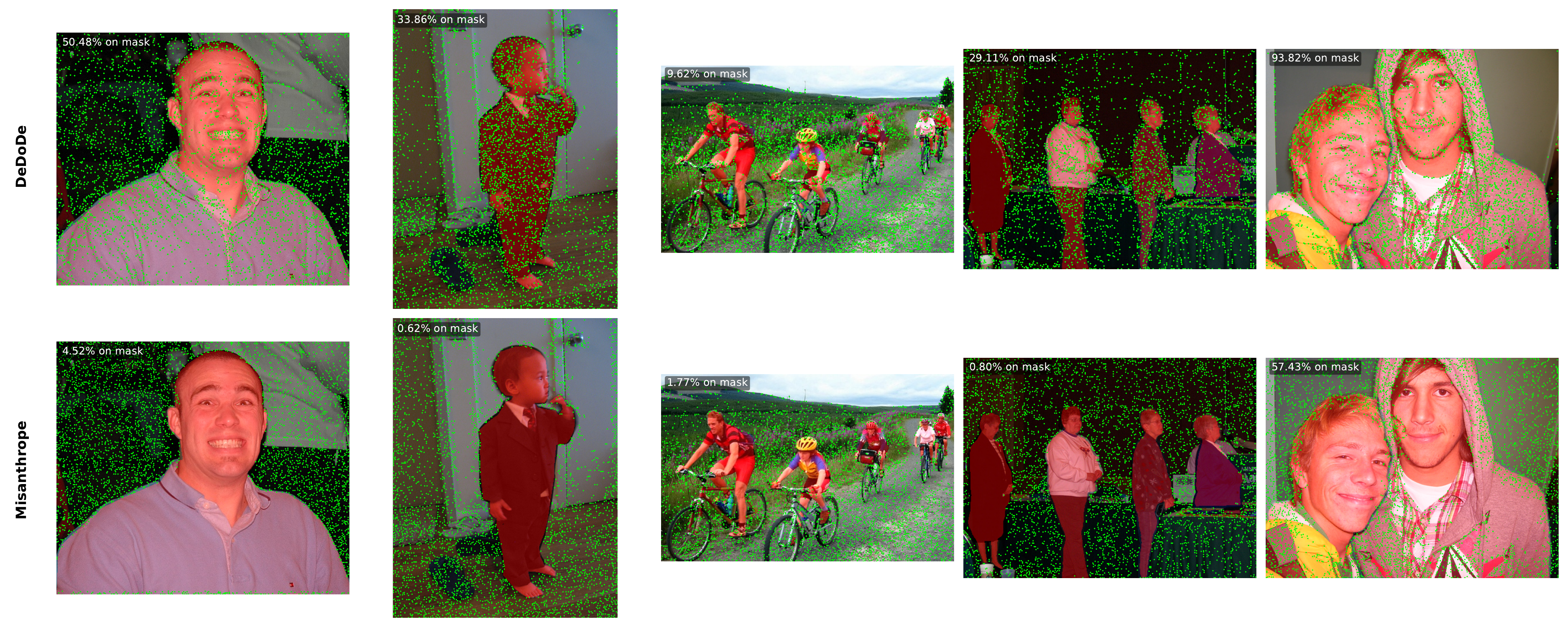}
    \caption{Examples from PascalVOC2012~\cite{pascalVoc} of $10{,}000$ keypoints per image detected by DeDoDe~\cite{dedode} and Misanthrope. Regions annotated as people are highlighted in red. Top row: DeDoDe; bottom row: Misanthrope.}
    \label{fig:VOCexamples}
    \vspace{-15pt}
\end{figure*}

\subsubsection{Person Avoidance in Keypoint Detection}
\label{subsubsec:personavoidance}
\begin{figure}
    \centering
    \includegraphics[width=0.7\linewidth]{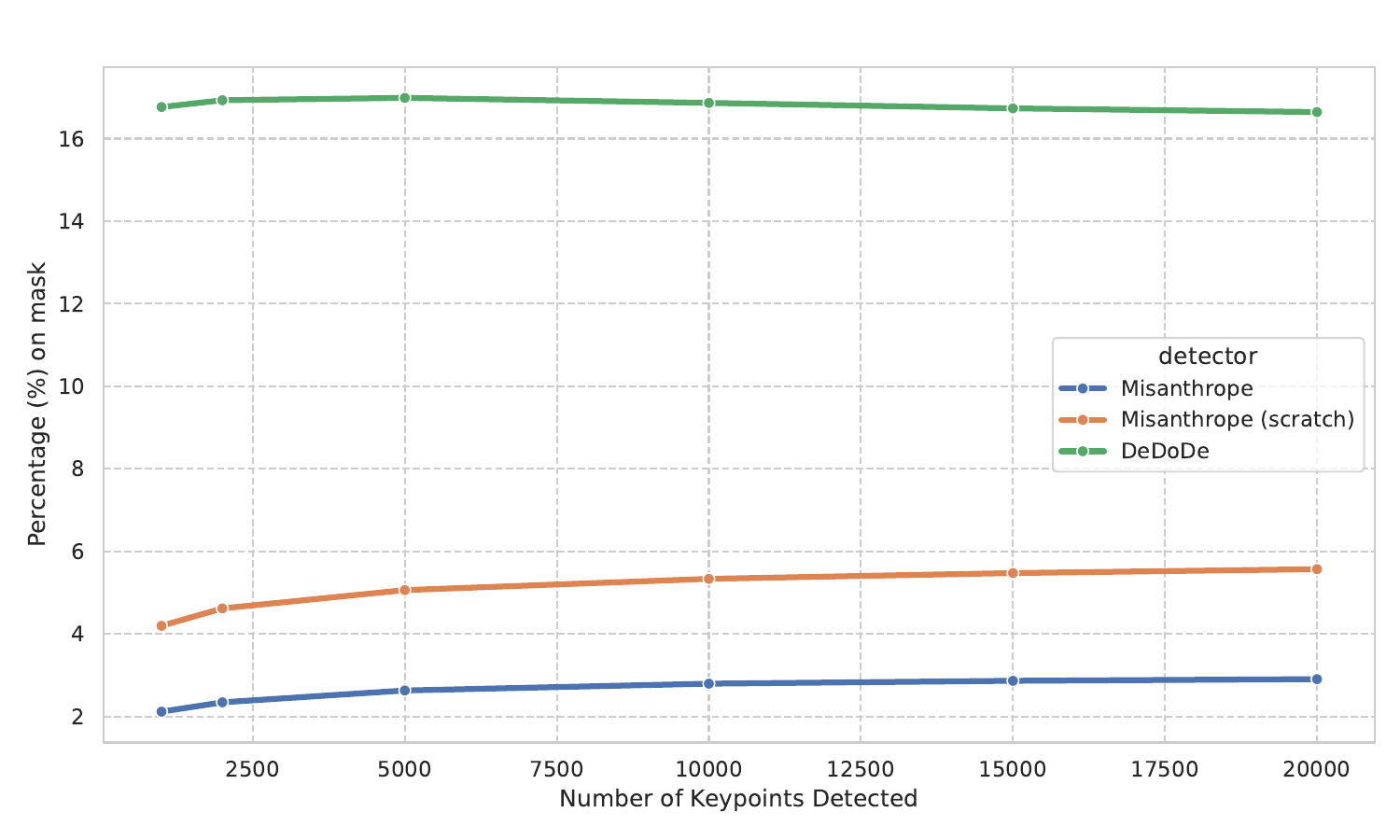}
    \caption{Percentage of keypoints falling within person-labeled regions on PascalVOC2012~\cite{pascalVoc}, as a function of the total number of extracted keypoints, for DeDoDe in green, Misanthrope in blue, and the ablation $\text{Misanthrope}_s$ (trained from scratch) in orange.}
    \label{fig:precision_accuracy}
\end{figure}
We begin by evaluating whether our keypoint detector effectively avoids detecting keypoints on people. For this experiment, we use the PascalVOC2012 dataset~\cite{pascalVoc}, filtering for images in which at least 1\% of pixels are labeled as \textit{person}, yielding 888 images drawn from both the training and validation splits. Figure~\ref{fig:VOCexamples} illustrates 10,000 keypoint detections from DeDoDe and Misanthrope. While DeDoDe frequently places keypoints on people, Misanthrope consistently avoids doing so. 

A notable failure case is shown on the far right of Figure~\ref{fig:VOCexamples}, where Misanthrope assigns approximately 60\% of its 10,000 extracted keypoints to person-labeled regions (compared to 94\% for DeDoDe on the same image). Such cases typically arise in portrait-like scenarios, where individuals dominate the frame against a plain or uniform background (e.g., a blank wall or open sky), leaving few alternative regions on which to place keypoints. Even in these situations, Misanthrope tends to avoid faces and instead places keypoints on clothing, suggesting it has learned a robust representation of person-like appearance.

Figure~\ref{fig:precision_accuracy} provides quantitative results across the filtered dataset for detection thresholds ranging from 1,000 to 20,000 keypoints. From the plot we can observe the percentage of keypoints detected on people to range from 2\% to 3\% for the Misanthrope detector, a more than fivefold reduction compared to the original DeDoDe~\cite{dedode} base detector. Importantly, while this percentage remains largely stable across thresholds, the absolute number of keypoints falling on person-labeled regions continues to grow proportionally with the extraction budget: at 20,000 keypoints, even 3\% corresponds to roughly 600 person keypoints, compared to approximately 30 at a threshold of 1,000. This distinction matters for privacy analysis: the near-flat curve reflects a consistent suppression \textit{rate}, but does not imply a ceiling on the total amount of person-information extracted.

Beyond keypoint locations themselves, modern local feature extraction networks implicitly encode appearance information from surrounding image regions in the descriptor associated with each keypoint, owing to the large receptive fields of the underlying networks. The privacy implications of this are difficult to characterize analytically; instead, we assess them empirically through person re-identification experiments, presented in the following subsection.
\vspace{-15pt}

\subsubsection{Person Detection and Re-Identification in Reconstructed Images}
\begin{figure}
    \centering
    \includegraphics[width=0.66\linewidth]{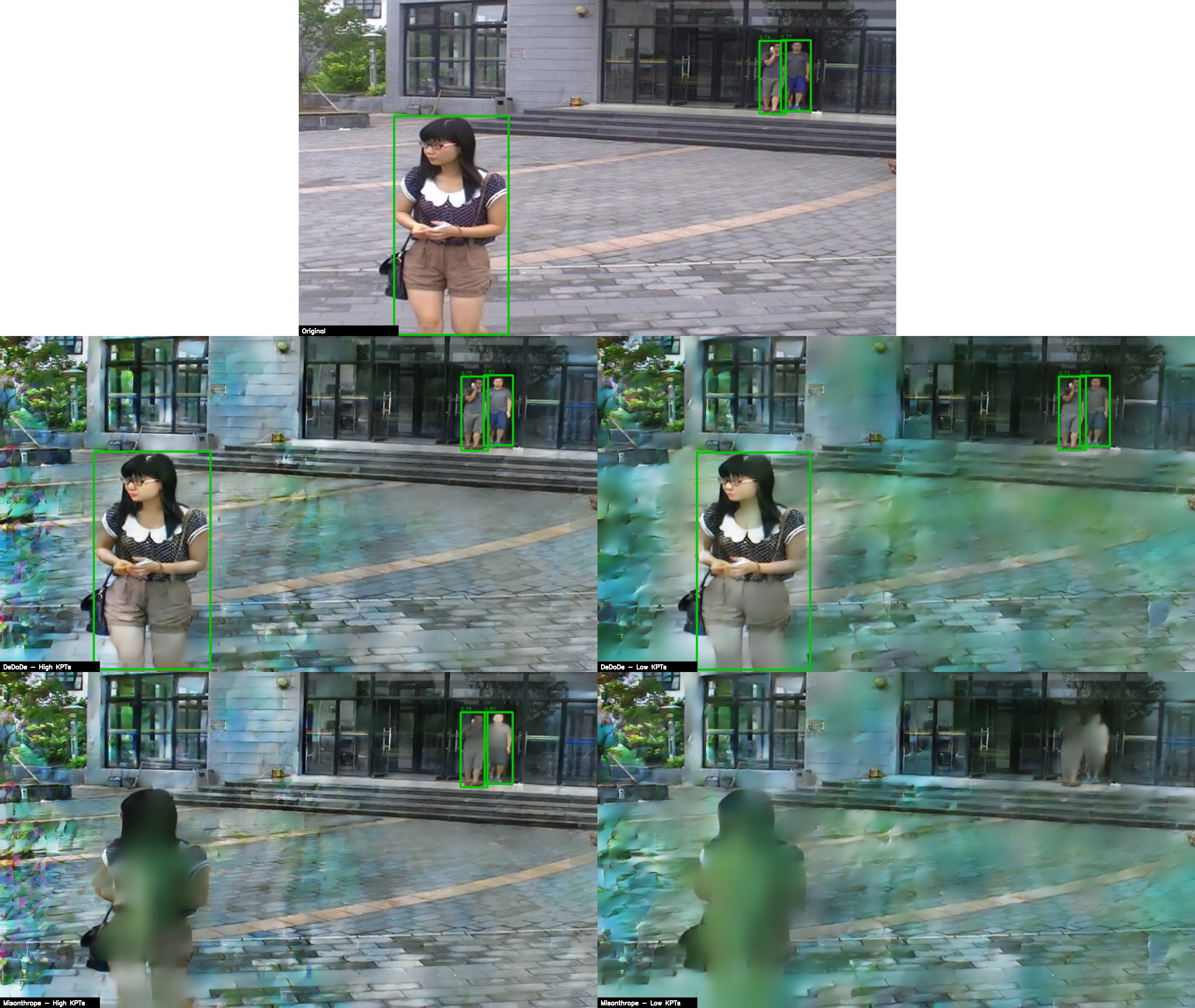}
    \caption{Examples of image inversion and person detection~\cite{yolov8_ultralytics} results on the PRW dataset~\cite{prw}. Top: original image. Middle row: inversions from DeDoDe~\cite{dedode} features; bottom row: inversions from Misanthrope features. Left: High Keypoint Threshold (HKT); right: Low Keypoint Threshold (LKT), based on thresholds from the IMC2021~\cite{IMC2020}.}
    \label{fig:inversion_examples}
\end{figure}
\label{subsubsec:persondetectreid}
To evaluate the privacy implications of feature inversion, we measure how effectively people can be detected and re-identified in images reconstructed from local features. We use the PRW dataset~\cite{prw}, which provides person bounding box annotations suitable for both detection and re-identification evaluation. For each image, we extract local features using DeDoDe and Misanthrope at two thresholds motivated by the Image
Matching Challenge 2021~\cite{IMC2020}: a Low Keypoint Threshold (LKT) of 0.27\% of pixels as keypoints and a High Keypoint Threshold (HKT) of 1.07\% of pixels as keypoints. We then reconstruct images from these features
using InvSfM~\cite{invsfm} and assess the privacy leakage of the
reconstructions.

\paragraph{Person detection.}

We run YOLOv8~\cite{yolov8_ultralytics} on the original images and on all reconstructions. Table~\ref{tab:person_detection_results} reports precision and recall. At LKT, Misanthrope reduces detection recall to 2.74\%, compared to 63.44\% for DeDoDe and 93.41\% on the original images; precision drops correspondingly to 8.86\%. At HKT, the gap narrows: Misanthrope's precision is comparable to DeDoDe's (47.28\% vs.\ 57.98\%), but recall remains substantially lower (58.68\% vs.\ 72.29\%). This is consistent with the trend observed in the Person Avoidance subsection: as the number of extracted keypoints increases, more keypoints inevitably fall on person regions, providing the inversion model with sufficient information to partially reconstruct them. Figure~\ref{fig:inversion_examples} illustrates this qualitatively: at LKT, Misanthrope's reconstructions render people as indistinct blurs that the detector largely fails to recognize, whereas DeDoDe's reconstructions preserve enough detail for reliable detection.

\begin{table}
    \centering
    \begin{tabular}{lccccc}
        \toprule
        \toprule
        & & \multicolumn{2}{c}{LKT} & \multicolumn{2}{c}{HKT} \\
        \cmidrule(lr){3-4} \cmidrule(lr){5-6}
        Metric & Original & DeDoDe~\cite{dedode} & Misanthrope & DeDoDe~\cite{dedode} & Misanthrope \\
        \midrule
        Precision & 59.75\% & 47.98\% & \textbf{8.86}\% & 57.98\% & \textbf{47.28}\% \\
        Recall & 93.41\% & 63.44\% & \textbf{2.74}\% & 72.29\% & \textbf{58.68}\% \\
        \bottomrule
    \end{tabular}
    \caption{Comparison of person detection metrics across feature extraction strategies in the PRW dataset~\cite{prw} using YOLOv8~\cite{yolov8_ultralytics}. LKT: Low Keypoint Threshold, HKT: High Keypoint Threshold. In \textbf{bold} the most privacy-preserving results.}
    \label{tab:person_detection_results}
\end{table}

\paragraph{Person re-identification.}

We also evaluate whether individuals can be re-identified from reconstructed images using OSNet~\cite{osnet}. We report Rank-1, Rank-5, and Rank-10 retrieval accuracy. We use both ground-truth bounding boxes and bounding boxes predicted by YOLO to detect people in the reconstructed images. The latter represents the more realistic threat scenario, as an attacker would not have access to ground-truth annotations.

\begin{table*}[t]
\centering
\resizebox{\textwidth}{!}{%
\begin{tabular}{@{}ccccccccccccc@{}}
\toprule \toprule
                       & \multicolumn{6}{c}{GT Bounding Boxes}                                         & \multicolumn{6}{c}{YOLO Bounding Boxes}                                       \\
\cmidrule(lr){2-7} \cmidrule(lr){8-13}
                       & \multicolumn{3}{c}{LKT}               & \multicolumn{3}{c}{HKT}               & \multicolumn{3}{c}{LKT}               & \multicolumn{3}{c}{HKT}               \\
\cmidrule(lr){2-4} \cmidrule(lr){5-7} \cmidrule(lr){8-10} \cmidrule(lr){11-13}
                       & \shortstack{R@1\\(\%)} & \shortstack{R@5\\(\%)} & \shortstack{R@10\\(\%)} & \shortstack{R@1\\(\%)} & \shortstack{R@5\\(\%)} & \shortstack{R@10\\(\%)} & \shortstack{R@1\\(\%)} & \shortstack{R@5\\(\%)} & \shortstack{R@10\\(\%)} & \shortstack{R@1\\(\%)} & \shortstack{R@5\\(\%)} & \shortstack{R@10\\(\%)} \\
\midrule
Original               & 61.89 & 75.26 & 80.31           & 61.89 & 75.26 & 80.31           & 54.35 & 72.24 & 77.20           & 54.35 & 72.24 & 77.20           \\
DeDoDe~\cite{dedode} & 28.25 & 45.75 & 54.06           & 54.21 & 71.56 & 77.44           & 14.75 & 33.12 & 43.43           & 38.81 & 62.60 & 70.60           \\

Misanthrope & \textbf{1.94} & \textbf{5.59} & \textbf{8.70} & \textbf{12.69} & \textbf{27.47} & \textbf{36.02} & \textbf{1.04} & \textbf{2.43} & \textbf{3.70} & \textbf{6.91} & \textbf{17.01} & \textbf{24.02} \\

\midrule
DeDoDe+SN            & 2.77  & 6.27  & 8.51            & 5.35  & 9.58  & 12.49           & 0.83  & 2.10  & 2.98            & 1.71  & 3.52  & 4.49            \\
$\text{Misanthrope}_s$ & 6.27  & 12.54 & 17.21           & 25.13 & 41.47 & 48.27           & 2.68  & 8.29  & 10.73           & 14.01 & 27.93 & 36.57           \\
\bottomrule
\end{tabular}%
}
\caption{Comparison of person re-identification metrics across feature detection strategies in the PRW dataset~\cite{prw} ($544$ identities, $19{,}116$ gallery images) using OSNet~\cite{osnet}. LKT: Low Keypoint Threshold, HKT: High Keypoint Threshold, GT: ground truth. In \textbf{bold} the most privacy-preserving results among the feature extraction strategies in the upper block. Below the midrule we report ablation and baseline strategies: weight initialization from scratch ($\text{Misanthrope}_s$) and filtering with a separate segmentation network (DeDoDe+SN).}
\label{tab:person_reid_results_both}
\end{table*}

As shown in Table~\ref{tab:person_reid_results_both}, at LKT with ground-truth boxes, Misanthrope reduces Rank-1 accuracy to 1.94\%, compared to 28.25\% for DeDoDe and 61.89\% on the original images. With YOLO-predicted boxes the results are even more favorable: Misanthrope achieves a Rank-1 of just 1.04\%, reflecting the compounding effect of the detector's inability to localize people in Misanthrope's reconstructions. At HKT the privacy advantage diminishes but remains meaningful: Misanthrope's Rank-1 with ground-truth boxes is 12.69\% versus 54.21\% for DeDoDe, and with YOLO boxes 6.91\% versus 38.81\%.

$\text{Misanthrope}_s$ consistently shows higher re-identification rates than the distilled Misanthrope across all settings. This reinforces the finding from Section~\ref{subsec:ablation}: the randomly initialized variant learns to detect good keypoints but fails to fully acquire the person-avoidance behavior, resulting in weaker privacy protection.

We also report results for a baseline strategy in which a separate segmentation network is used to filter out keypoints on people (DeDoDe+SN). At LKT with ground-truth boxes, this approach achieves a Rank-1 of 2.77\%, slightly higher than Misanthrope's 1.94\%. At HKT, DeDoDe+SN outperforms Misanthrope with ground-truth boxes (5.35\% vs.\ 12.69\%) and YOLO bounding boxes (1.71\% vs.\ 6.91\%). Thus, at higher keypoint thresholds the separate segmentation network filters person keypoints more effectively, while at lower thresholds Misanthrope's integrated approach is competitive and in some cases superior. Overall, Misanthrope embeds person avoidance directly into the keypoint detection process: since its architecture is unchanged, its inference cost is identical to the base detector's by construction, whereas DeDoDe+SN requires an additional segmentation forward pass (YOLOv8~\cite{yolov8_ultralytics} in this experiment).

Taken together, these results demonstrate that Misanthrope's suppression of keypoints on people translates directly into reduced privacy leakage under feature inversion. At the low keypoint threshold typical of resource-constrained deployments, person detection recall from Misanthrope's features drops to under 3\%, and re-identification accuracy is dramatically reduced compared to both the original images and DeDoDe reconstructions. While not fully at chance (a random ranking baseline yields approximately 0.16\% Rank-1 given the PRW gallery size of $19{,}116$ images), the residual re-identification performance is sufficiently low to substantially mitigate the practical threat posed by inversion attacks.

\subsection{Matching Performance}
\label{subsec:matching}
We evaluate the matching performance of Misanthrope on three benchmarks: HPatches~\cite{hpatches} (Table~\ref{tab:HPatches}) as an example of a well-established image matching dataset where no humans are present; Wild-SLAM~\cite{wildgs} (Table~\ref{tab:wild_results}) as an example of indoor scenes with humans as distractors; and the Image Matching Challenge 2021~\cite{IMC2020} Phototourism test set (Table~\ref{tab:maa_10}) as an example of challenging wide-baseline image matching with human distractors. 
In all our matching tests we use brute-force matching with cross-check and MAGSAC++~\cite{magsac} as our model estimator.

\subsubsection{Image Matching without Humans}

\begin{table}
\centering
\setlength{\tabcolsep}{3pt} 
\begin{tabular}{lcccccc}
\toprule
\toprule
Method & \multicolumn{3}{c}{MMA} & \multicolumn{3}{c}{MHA} \\
\cmidrule(lr){2-4} \cmidrule(lr){5-7}
& 1px & 3px & 5px & 1px & 3px & 5px \\
& [\%] & [\%] & [\%] & [\%] & [\%] & [\%] \\
\midrule
\cite{aliked}       ALIKED &\textbf{44.03}& \underline{74.88} & 82.71 & 52.93 & \textbf{84.14} &\textbf{90.00} \\
\cite{disk}         DISK & \underline{38.09} & \textbf{76.82}& \textbf{84.47} & 50.69 & 80.34 & 88.28 \\
\cite{dedode}       DeDoDe & 33.49 & 73.69 & \underline{83.24} & \textbf{56.90} & \underline{83.97} & \textbf{90.00}\\
\cite{superpoint}   SuperPoint & 25.83 & 61.02 & 72.14 & 50.69 & 81.55 & 89.14 \\
\cite{xfeat}        XFeat & 19.89 & 56.56 & 71.13 & 47.07 & 81.72 & \underline{89.83} \\
Misanthrope & 33.17 & 73.05 & 82.64 & \underline{54.31} & 81.72 & 88.79 \\
\midrule
$\text{Misanthrope}_s$ & 33.84 & 73.49 & 82.91 & 55.86& 82.07 & 89.83 \\
\bottomrule
\end{tabular}
\caption{Mean Matching Accuracy (MMA) and Mean Homography Accuracy (MHA) on HPatches~\cite{hpatches} at 1, 3, and 5 pixel thresholds. In \textbf{bold} the best method; \underline{underlined} the second best.}
\label{tab:HPatches}
\end{table}
On HPatches we follow the evaluation procedure of~\cite{alike}. For each method we compute the Mean Matching Accuracy (\textbf{MMA}), the fraction of putative matches that are inliers under the ground-truth homography, and the Mean Homography Accuracy (\textbf{MHA}), the fraction of image pairs for which the estimated homography projects the image corners within the given pixel threshold of their ground-truth locations. The former measures the accuracy of the raw matches; the latter, the accuracy of the full homography estimation pipeline. 
Looking at Table~\ref{tab:HPatches} we can observe that the performance of Misanthrope is slightly lower than, but in line with, that of the original DeDoDe model. Across all thresholds, Misanthrope stays within $0.7$ percentage points of DeDoDe on MMA and within $2.6$ points on MHA. This suggests that the proposed self-distillation did not negatively affect the keypoint detection abilities of the model in scenes where no people are present. 

\subsubsection{Image Matching with Human Distractors}
\begin{table*}
\centering
\small 
\setlength{\tabcolsep}{2pt}
{
\begin{tabularx}{\textwidth}{l @{\extracolsep{\fill}} cccccccccc} 
\toprule
\toprule
\makecell{Scene} & \makecell{ 1} & \makecell{2} & \makecell{3} & \makecell{4} & \makecell{5} & \makecell{6} & \makecell{7} & \makecell{8} & \makecell{9} & \makecell{Avg. \\ Rank} \\

\midrule
\cite{aliked}    ALIKED  & 72.69 & 71.24 & \underline{82.87} & 50.79 & 82.24 & \underline{37.47} & 57.73 & 75.86 & 78.83 & 2.78  \\
\cite{disk}      DISK  & 68.53 & 71.07 & 78.46 & 48.22 & 81.37 & \textbf{43.03} & 56.38 & 74.83 & 74.53 & 3.56 \\
\cite{dedode}    DeDoDe  & \underline{82.36} & \underline{77.87} & \textbf{84.14} & \underline{52.59} & \underline{87.11} & 33.47 & \underline{63.21} & \underline{81.31} & \underline{88.87} & 2.00 \\
\cite{superpoint} SuperPoint & 54.24 & 53.73 & 72.48 & 37.74 & 58.00 & 26.34 & 34.34 & 61.74 & 53.54 & 5.56 \\
\cite{xfeat}     XFeat  & 61.52 & 61.46 & 71.27 & 40.09 & 63.39 & 23.39 & 47.29 & 56.04 & 63.62 & 5.22  \\
Misanthrope  & \textbf{83.25} & \textbf{78.08} & 67.35 & \textbf{53.00} & \textbf{87.41} & 28.18 & \textbf{64.13} & \textbf{81.62} & \textbf{89.01} & \textbf{1.89}  \\
\midrule
$\text{Misanthrope}_s$ & 83.39 & 78.39 & 75.54 & 53.41 & 87.07 & 29.78 & 64.45 & 81.27 & 89.15 & - \\
\bottomrule
\end{tabularx}
}
\caption{Mean Average Accuracy (mAA) at $10^\circ$ on the IMC2021~\cite{IMC2020} Phototourism test set. Scenes: 1 British Museum, 2 Florence Cathedral Side, 3 Lincoln Memorial Statue, 4 London Bridge, 5 Milan Cathedral, 6 Mount Rushmore, 7 Piazza San Marco, 8 Sagrada Familia, 9 St.\ Paul's Cathedral. In \textbf{bold} the best method per scene; \underline{underlined} the second best; $\text{Misanthrope}_s$ (ablation) is excluded from the ranking.}
\label{tab:maa_10}
\end{table*}

We also evaluate the model in scenarios in which humans can act as a distractor for the image matching process. To this end, the first dataset we consider is the Phototourism test set from the Image Matching Challenge 2021~\cite{IMC2020}. The dataset consists of nine scenes, each comprising 100 posed images calibrated through SfM; the poses are defined up to an unknown scale factor.
The challenge requires matching each possible image pair, computing the essential matrix, and from it the relative transformation. We follow the original challenge requirements~\cite{IMC2020} and extract up to 8000 keypoints per image, scoring the estimated transformations using the mean Average Accuracy (\textbf{mAA}) at 10 degrees. The mAA is computed by taking the maximum between the rotation and translation angular distances for each image pair and then computing for each scene the percentage of image pairs whose angular error is below 10 degrees. The rotational and translational angular distances are the angles between the rotation matrices and between the normalized translation vectors, respectively. As can be seen in Table~\ref{tab:maa_10}, Misanthrope outperforms all other models across all scenes except the Lincoln Memorial and Mount Rushmore (scenes 3 and 6). In both cases, the primary subjects are statues of people, which the model still interprets as people and consequently avoids detecting keypoints on, leading to poor results.

\begin{figure}
    \centering
    \includegraphics[width=0.67\linewidth]{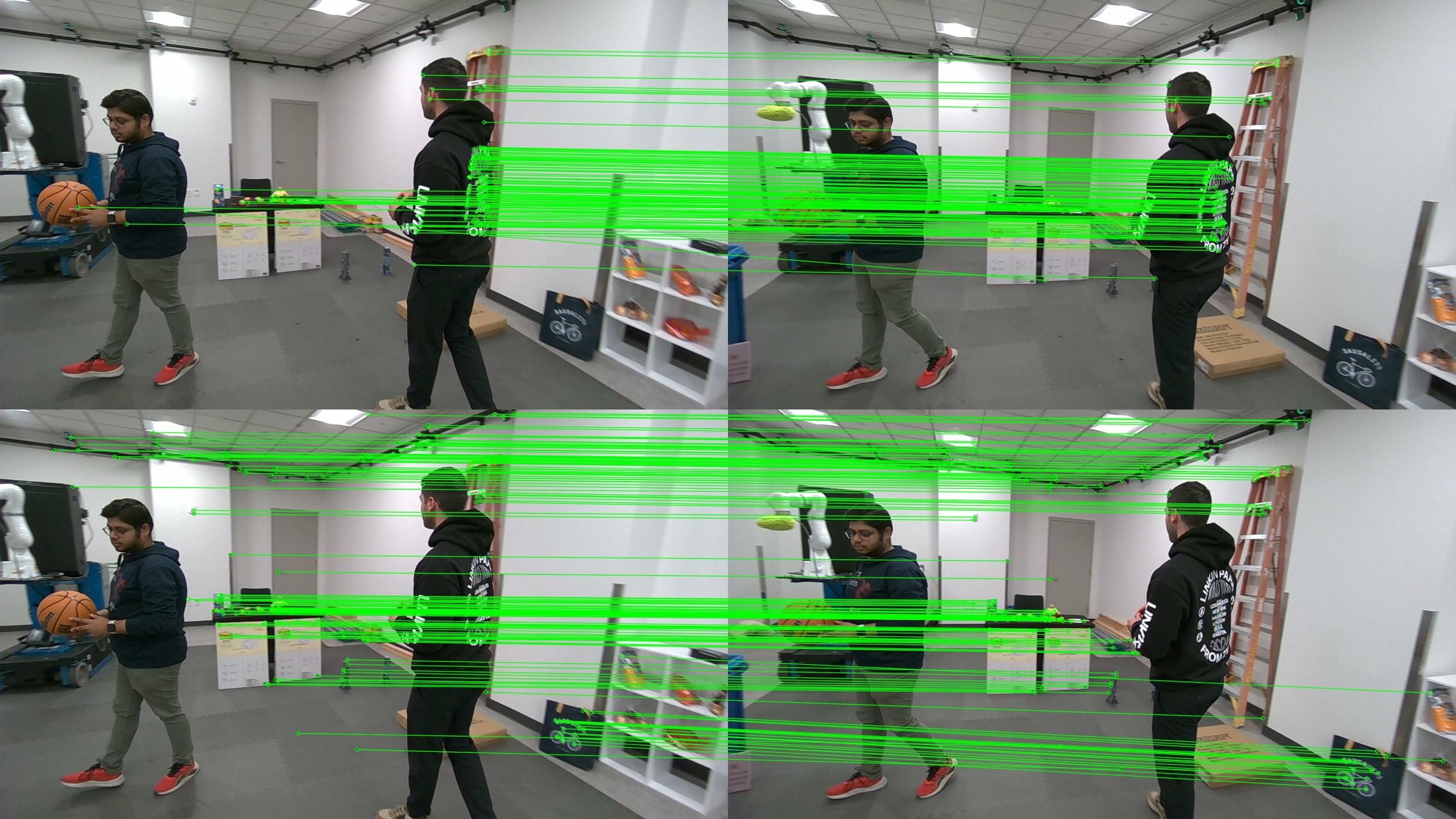}
    \caption{Image matching example from the Wild-SLAM dataset~\cite{wildgs}. In green, the inliers of the essential matrix estimation according to MAGSAC++~\cite{magsac}. Top: DeDoDe~\cite{dedode}; bottom: Misanthrope (ours). Misanthrope avoids detecting keypoints on people, yielding a more accurate model estimate.}
    \label{fig:slamdunk}
\end{figure}
We then evaluate Misanthrope's performance on an indoor dataset, Wild-SLAM~\cite{wildgs} (Table~\ref{tab:wild_results}). This SLAM dataset contains 30-frames-per-second image sequences with ground truth from a motion tracking system. To evaluate the image matching methods we select image pairs 30 frames apart and discard all image pairs whose distance is less than 10 centimeters to avoid degenerate translation estimates. For each image pair 2000 keypoints are extracted, and the matches from each method are used to compute the essential matrix, from which the relative transformation is recovered; we then compute the mAA as for the IMC2021~\cite{IMC2020} data. Misanthrope yields a 2\% and a 12\% improvement in median rotation and translation error, respectively, relative to the best-performing baseline. We attribute these improvements to more matches focusing on the static parts of the scene rather than on people moving through it. An example of poor model estimation due to matches on dynamic objects is visible in Figure~\ref{fig:slamdunk}.

\begin{table}
\centering
\begin{tabular}{lccccc}
\toprule
\toprule
 Method & MRE & MTE & \multicolumn{3}{c}{mAA} \\
   &  [$^\circ$]&  [$^\circ$] & @$1^\circ$ & @$5^\circ$ & @$10^\circ$\\
\midrule

  \cite{aliked}    ALIKED & \underline{0.43} & 1.39 & 35.02 & 85.95 & \underline{93.14} \\
  \cite{dedode}    DeDoDe & \underline{0.43} & \underline{1.38} & \underline{35.31} & 83.12 & 88.98 \\
  \cite{disk}      DISK & 0.45 & 1.56 & 31.65 & 83.12 & 91.91 \\
  \cite{superpoint} SuperPoint & \underline{0.43} & 1.53 & 31.34 & \underline{86.20} & \textbf{93.94} \\
  \cite{xfeat}     XFeat & 0.46 & 1.82 & 27.85 & 79.97 & 88.99 \\
  Misanthrope & \textbf{0.42} & \textbf{1.21} & \textbf{39.94}& \textbf{87.21} & 91.45 \\
  \midrule
  $\text{Misanthrope}_s$& 0.41 & 1.18 & 41.01 & 87.85 & 91.70 \\
\bottomrule
\end{tabular}
\caption{Relative pose estimation results on the Wild-SLAM dataset~\cite{wildgs}; Median Rotational Error (MRE), Median Translational Error (MTE), mean Average Accuracy (mAA) at 1, 5, and 10 degrees. In \textbf{bold} the best model; \underline{underlined} the second best.}
\label{tab:wild_results}
\end{table}

\subsection{Ablation Experiments}
\label{subsec:ablation}
To disentangle the contribution of the self-distillation from the training data, we train a variant, $\text{Misanthrope}_s$, in which the student's weights are randomly initialized rather than initialized from the pre-trained teacher; all other training parameters remain unchanged. As shown in Table~\ref{tab:HPatches} and Table~\ref{tab:maa_10}, $\text{Misanthrope}_s$ achieves matching performance on par with or slightly above the distilled model. However, its ability to avoid detecting keypoints on people is substantially degraded (Figure~\ref{fig:precision_accuracy}). We attribute this to the difference in learning burden: the distilled model inherits weights that already encode what constitutes a good keypoint, and needs only to acquire the additional constraint of person avoidance. The randomly initialized model must learn both tasks simultaneously from the same filtered supervision, and the results suggest that the person-avoidance behavior is the harder of the two to acquire under these conditions. Interestingly, $\text{Misanthrope}_s$ achieves marginally better matching performance than the distilled variant; we conjecture that this is because the distilled model inherits a prior that favors keypoint-rich regions on people, creating a residual tension with the person-avoidance objective that may subtly affect detection quality elsewhere.
\vspace{-5pt}
\section{Conclusion}
\label{sec:conclusion}
\vspace{-10pt}
We have presented Misanthrope, a privacy-aware keypoint detector trained via self-distillation to avoid detecting features on people. By integrating semantic avoidance directly into the detector rather than relying on post-hoc obfuscation, our approach suppresses privacy-sensitive data at the source: information that is never extracted cannot be recovered by future inversion algorithms. 
Experiments on feature inversion demonstrate that Misanthrope substantially reduces person detection and re-identification from reconstructed images. Crucially, this privacy benefit does not come at the expense of matching performance: in scenes containing people, Misanthrope matches or outperforms existing detectors by avoiding spurious correspondences on human regions. 
Looking ahead, the framework can in principle be applied to any keypoint detector and extended to suppress arbitrary semantic categories. We believe source-level privacy preservation is a necessary component of any distributed vision system that shares potentially sensitive visual data with third parties.

\vspace{-5pt}
\section*{Acknowledgments}
This work is part of the EU-Horizon project \href{https://www.egeniouss.eu/}{\textit{egeniouss}}, which received funding under the call HORIZON-EUSPA-2021-Space with the project number 101082128.
%
%
\bibliographystyle{splncs04}
\bibliography{main}
\end{document}